%% file: main.tex
\documentclass[11pt]{article}

\usepackage[preprint]{acl}

\usepackage{times}
\usepackage{latexsym}
\usepackage{booktabs}   
\usepackage{tabularx}   
\usepackage{siunitx}    
\usepackage{makecell}   
\usepackage{threeparttable} 
\usepackage{url}        

\usepackage{multirow}
\usepackage{algorithm}
\usepackage{algpseudocode}
\usepackage{amsmath,amssymb}
\usepackage[T1]{fontenc}

\usepackage[utf8]{inputenc}

\usepackage{microtype}
\usepackage{amsmath}
\usepackage{longtable}
\usepackage[most]{tcolorbox}

\usepackage{inconsolata}

\usepackage{graphicx}

\title{AgentDevel: Reframing Self-Evolving LLM Agents as Release Engineering}

\author{Di Zhang \\
  Fudan University \\
  \texttt{di.zhang@ustc.edu} \\}

\begin{document}
\maketitle
\begin{abstract}
Recent progress in large language model (LLM) agents has largely focused on embedding self-improvement mechanisms inside the agent or searching over many concurrent variants. While these approaches can raise aggregate scores, they often yield unstable and hard-to-audit improvement trajectories, making it difficult to guarantee non-regression or to reason about failures across versions. We reframe agent improvement as \textbf{release engineering}: agents are treated as shippable artifacts, and improvement is externalized into a regression-aware release pipeline.
We introduce \textbf{AgentDevel}, a release engineering pipeline that iteratively runs the current agent, produces implementation-blind, symptom-level quality signals from execution traces, synthesizes a single release candidate (RC) via executable diagnosis, and promotes it under flip-centered gating. AgentDevel features three core designs: (i) an implementation-blind LLM critic that characterizes failure appearances without accessing agent internals, (ii) script-based executable diagnosis that aggregates dominant symptom patterns and produces auditable engineering specifications, and (iii) flip-centered gating that prioritizes pass→fail regressions and fail→pass fixes as first-class evidence.
Unlike population-based search or in-agent self-refinement, AgentDevel maintains a single canonical version line and emphasizes non-regression as a primary objective. Experiments on execution-heavy benchmarks demonstrate that AgentDevel yields stable improvements with significantly fewer regressions while producing reproducible, auditable artifacts. Overall, AgentDevel provides a practical development discipline for building, debugging, and releasing LLM agents as software development.

\end{abstract}

\section{Introduction}

Large language model (LLM)–based agents are no longer just chatbots. Today, many agents can take actions in the world: they browse websites, call external tools and APIs, write code, and solve tasks that require multiple steps and decisions. For agents deployed in production workflows, unreliable improvement can be more damaging than no improvement at all. As these agents move closer to real applications, we face a very practical question: how do we improve an agent reliably over time when it repeatedly fails?

A natural idea is to let the agent “improve itself.” Recent work follows this direction in several ways. Some approaches ask the agent to reflect on feedback and store reflections in memory so it can do better next time (Reflexion~\citep{shinn2023reflexionlanguageagentsverbal}). Others iterate a loop where the model critiques its own output and rewrites it (Self-Refine~\cite{madaan2023selfrefineiterativerefinementselffeedback}). Still others treat prompts as objects to be searched or evolved (PromptBreeder~\cite{fernando2023promptbreederselfreferentialselfimprovementprompt}). These methods can indeed raise benchmark scores, but they often come with a familiar frustration for anyone who has built real systems: improvements can feel brittle and difficult to control. After a change, it may be unclear whether we introduced a regression, whether the improvement can be reproduced, or why certain failures suddenly appear. In other words, we may get better averages while losing confidence in the system’s stability.

This paper argues that the root issue is partly a matter of framing. Many self-improvement methods treat improvement as something that lives \textit{inside the agent}—as if the agent should become a self-evolving organism. We take a different view: in real software development, systems rarely improve because the program “reflects.” They improve because developers build an external workflow around them: we collect logs, run tests, diagnose failures, and release new versions only after they pass checks. If you have used continuous integration (CI), you already know a key lesson: average performance is not enough. What often matters most is how \textit{individual test cases} change across versions. In particular, we pay close attention to pass→fail (P→F) flips (a previously working case now breaks, i.e., a regression) and fail→pass (F→P) flips (a previously failing case now works, i.e., a fix). This “flip” view gives us a concrete and intuitive way to think about agent improvement: we should improve an agent like we improve software—by controlling regressions and making progress visible and auditable.

Based on this thesis, we propose AgentDevel, a factory-style development pipeline that reframes agent improvement as release engineering. AgentDevel maintains a single canonical agent line, rather than creating many competing variants. Each iteration produces exactly one release candidate (RC); only if the RC passes acceptance checks is it promoted to the next official version. To start an iteration, we run the current agent on a development set (TrainSet) and record structured execution traces: what actions the agent took, which tools it called, what it observed, what errors occurred, and what final output it produced. When available, we also run programmatic scorers—non-LLM checks such as unit tests, schema validation, and format checking—to obtain deterministic pass/fail signals.

Next, we introduce an independent LLM critic to provide higher-level, human-like quality characterization. The critic is intentionally designed to be implementation-blind: it does not see the agent’s internal design, which we refer to as the agent’s blueprint—the agent’s prompt, code, and tooling internals (i.e., how it is constructed). Instead, the critic only receives (i) the rubric (what counts as correct), (ii) the execution traces, and (iii) optionally the programmatic scoring results. Importantly, the critic does not perform causal attribution and does not propose repairs. Its job is simply to describe what went wrong at the surface level and to group failures into symptom-like categories such as “missing a required step,” “wrong order of actions,” or “invalid tool arguments.” The set of symptom categories is not fixed; it can evolve over time as new failure appearances emerge.

With traces and symptom descriptions in hand, AgentDevel then generates and executes diagnostic scripts (e.g., Python) that summarize dominant failure appearances, typical triggering conditions, representative examples, and how frequently each issue occurs. These scripts are regenerated each iteration, using the previous iteration’s scripts as soft references, which enables a bootstrapped diagnosis process rather than a rigid template system. Based on this diagnosis, AgentDevel synthesizes one RC that may modify prompts, code, or tool wrappers in the blueprint. The RC is then evaluated on the same TrainSet under a flip-centered gate: instead of focusing only on aggregate metrics, we emphasize example-level flips, especially P→F (regressions) and F→P (fixes). In release engineering terms, P→F cases are the most alarming—because they correspond to “breaking something that used to work.” RCs are promoted only when improvements are concentrated on the targeted symptom categories and unacceptable P→F regressions are limited. Iteration continues until marginal gains are exhausted and further changes begin to show clear overfitting signals; only then do we evaluate the final version once on a held-out TestSet.

AgentDevel is designed to be benchmark-agnostic: the same release workflow can be applied to domains with very different failure surfaces, including verifiable software engineering tasks, long-horizon web interaction, and tool/API composition. Across such regimes, our goal is to show that a single externalized, release-driven pipeline can produce agent improvements that are more stable, easier to audit, and less regression-prone than approaches that place self-improvement inside the agent or rely on multi-variant evolutionary search.

Contributions.

\begin{itemize}
    \item We introduce AgentDevel, a release-engineering paradigm and pipeline that externalizes agent improvement into a single canonical development line with RC-based promotion.
    \item We formalize an implementation-blind, symptom-level critic that consumes only rubrics, traces, and optional programmatic scoring results—separating surface characterization from causal diagnosis and repair.
    \item We propose flip-centered gating (P→F/F→P) together with executable diagnostic scripting for regression-aware, auditable agent improvement and principled stopping based on diminishing returns and overfitting risk.
\end{itemize}

\section{Method}
\begin{figure*}
    \centering
    \includegraphics[width=1.0\linewidth]{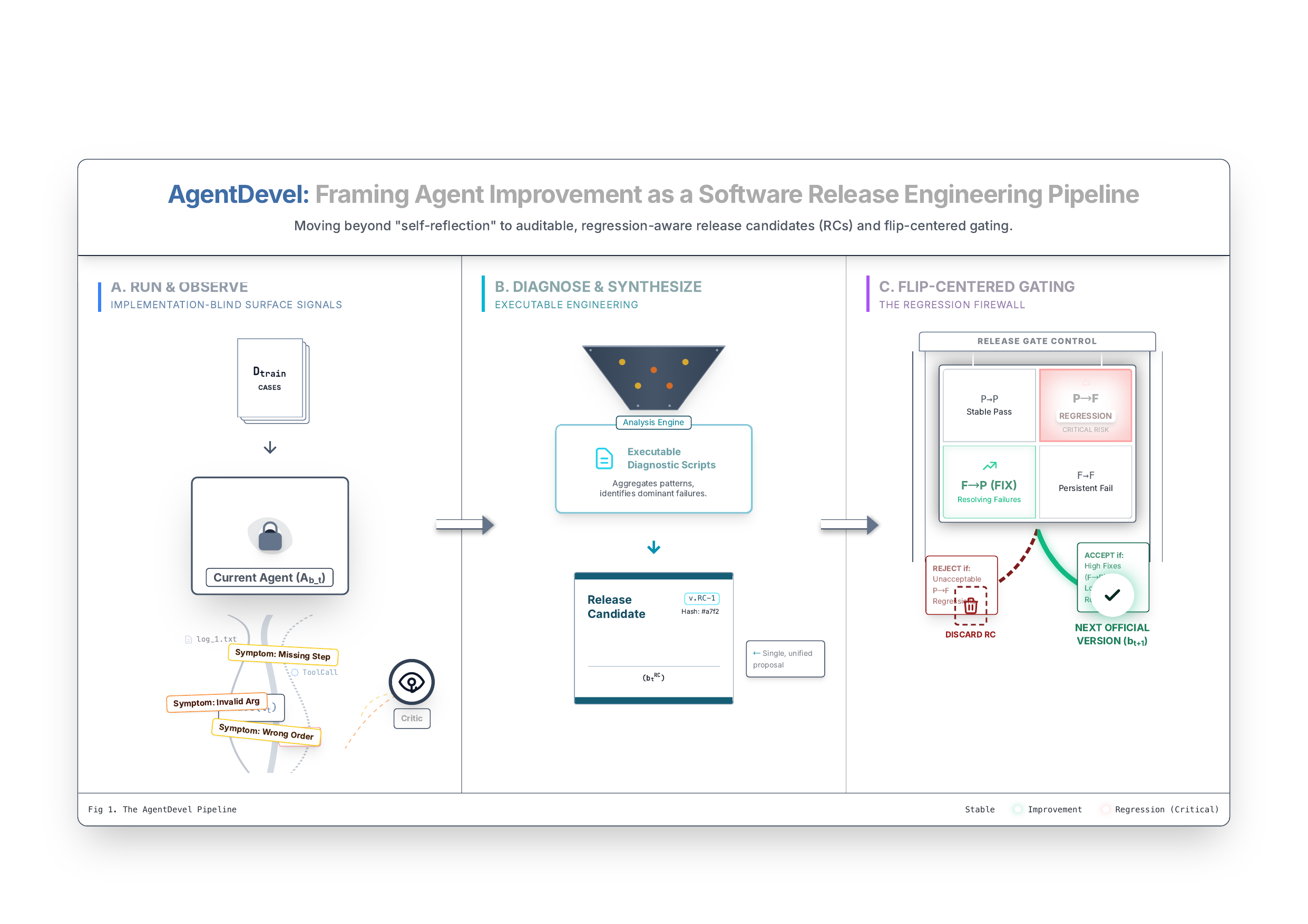}
    \caption{Main pipeline of AgentDevel}
    \label{fig:main}
\end{figure*}

\subsection{Setup \& Overview (What we build)}
\label{sec:setup}

We treat an LLM agent as a maintainable software artifact rather than a self-evolving organism. 
Concretely, we refer to the agent's internal design---its \emph{prompt}, \emph{code}, and \emph{tooling internals}---as the \textbf{agent blueprint}. 
Throughout the paper, ``improving an agent'' means proposing and validating changes to this blueprint.

\paragraph{Objects and data.}
We assume a task $\mathcal{T}$ comes with a development set and a held-out test set:
\[
D_{\text{train}} = \{x_i\}_{i=1}^{N}, \qquad 
D_{\text{test}}  = \{x_j\}_{j=1}^{M}.
\]
All iterative development is performed on $D_{\text{train}}$, while $D_{\text{test}}$ is used only once for final post-hoc reporting.

A blueprint $b$ induces a concrete agent $A_b$. 
Running the agent on an input $x$ produces both an output and a structured execution record:
\[
(\hat{y}, \tau) \leftarrow A_b(x),
\]
where $\tau$ is a \textbf{trace} that records execution steps such as actions, tool calls, observations, errors, and the final output. 
We use traces in the same spirit as observability in software systems: they capture what happened during an execution and support downstream diagnosis and analysis.

To evaluate behavior, we assume a \textbf{rubric} $R$ that specifies grading criteria. 
When available, we also use deterministic \textbf{programmatic scorers} (e.g., unit tests, schema or format checks), denoted by
\[
s(\hat{y}, \tau; R) \rightarrow g,
\]
where $g$ is a structured grading signal (e.g., pass/fail and/or an error code).

\paragraph{Single canonical line and release candidates.}
AgentDevel maintains a \emph{single canonical version line} of the agent.
At iteration $t$, the current blueprint is $b_t$ (with agent $A_{b_t}$). 
The pipeline produces exactly one \textbf{release candidate (RC)} blueprint $b_t^{\text{RC}}$.
If the RC passes gating, it is promoted to the next official version $b_{t+1}$; otherwise, it is discarded.

Figure~\ref{fig:main} provides a one-pass view of the pipeline:
\text{Run} $\rightarrow$ \text{Score} $\rightarrow$ \text{Critic} $\rightarrow$ \text{Diagnose(script)} $\rightarrow$ \text{RC} $\rightarrow$ \text{Gate(P$\rightarrow$F / F$\rightarrow$P)} $\rightarrow$ \text{Promote}.

\paragraph{Why we care about flips.}
A key lesson from regression testing is that after a change, we must ensure we did not break previously working behavior.
Let $p_t(x)\in\{0,1\}$ indicate whether example $x$ passes under blueprint $b_t$.
Comparing the current version $t$ and a candidate $t{+}1$, we define
\begin{align}
\mathrm{P2F}_t = \{x \in D_{\text{train}} : p_t(x)=1,\ p_{t+1}(x)=0\}
\\
\mathrm{F2P}_t = \{x \in D_{\text{train}} : p_t(x)=0,\ p_{t+1}(x)=1\}.
\end{align}
$\mathrm{P2F}_t$ corresponds to \textbf{pass$\rightarrow$fail} flips (regressions), while $\mathrm{F2P}_t$ corresponds to \textbf{fail$\rightarrow$pass} flips (fixes).
These flips form the core lens for gating decisions in AgentDevel.

\subsection{Running the Agent \& Producing Quality Signals (What We Observe)}
\label{sec:running}

Each AgentDevel iteration begins by running the current build and converting its behavior into auditable quality signals. At iteration $t$, we have an agent blueprint $b_t$ and its instantiated agent $A_{b_t}$. For every development case $x \in D_{\text{train}}$, we execute the agent and collect both the produced output and a structured execution trace:
\begin{equation}
(\hat{y}_t(x),\ \tau_t(x)) \leftarrow A_{b_t}(x).
\end{equation}
Here $\tau_t(x)$ is a trace that records the step-by-step execution (e.g., actions, tool calls, observations, errors, and the final output), serving as an observability artifact for downstream diagnosis.

\paragraph{Programmatic scoring (hard signals).}
When available, AgentDevel first applies deterministic, non-LLM programmatic scorers such as unit tests or schema/format validators. We denote a generic scorer as
\begin{equation}
g_t(x) \leftarrow s\!\left(\hat{y}_t(x),\ \tau_t(x);\ R\right),
\end{equation}
where $R$ is the rubric and $g_t(x)$ is a structured grading result (e.g., pass/fail and error codes). These signals are reproducible and provide hard, objective checks.

\paragraph{Implementation-blind LLM critic (soft signals).}
We then introduce an implementation-blind LLM critic that observes only the rubric $R$, the trace $\tau_t(x)$, and (optionally) the programmatic scorer output $g_t(x)$:
\begin{equation}
(\tilde{p}_t(x),\ \ell_t(x),\ d_t(x)) \leftarrow c\!\left(R,\ \tau_t(x),\ g_t(x)\right).
\end{equation}
The critic returns a pass/fail judgment $\tilde{p}_t(x)\in\{0,1\}$, a symptom label $\ell_t(x)$, and a short symptom description $d_t(x)$. We assume $c$ is deterministic given its inputs (or evaluated with a fixed random seed). The critic does not observe the agent blueprint $b_t$ (prompt, code, or tooling internals), performs no causal attribution, and proposes no repairs; it is restricted to symptom-level characterization.  This separation mirrors standard quality inspection practices in manufacturing, where inspectors report observable defects rather than hypothesized root causes.

\paragraph{Final pass indicator.}
Because deterministic checks are preferred when available, we define the final per-example pass indicator as
\begin{equation}
p_t(x)=
\begin{cases}
\mathrm{Pass}\!\left(g_t(x)\right), & \text{if $g_t$ available},\\[4pt]
\tilde{p}_t(x), & \text{otherwise}.
\end{cases}
\end{equation}

\paragraph{Open-ended symptom space.}
Symptom labels are task-local and allowed to expand over iterations:
\begin{equation}
\ell_t(x)\in\mathcal{L}_t,\qquad \mathcal{L}_t \subseteq \mathcal{L}_{t+1}.
\end{equation}

\paragraph{Per-example quality record.}
Each example yields an implementation-agnostic record
\begin{equation}
\small
r_t(x)=\\\big(\hat{y}_t(x),\ \tau_t(x),\ g_t(x),\ \tilde{p}_t(x),\ p_t(x),\ \ell_t(x),\ d_t(x)\big),
\end{equation}
where $g_t(x)$ may be absent if no programmatic scorer exists. These records constitute the quality signals consumed by the executable diagnosis stage. These records are the sole inputs to the executable diagnosis stage.

\subsection{Executable Diagnosis \& RC Synthesis (How We Propose Changes)}
\label{sec:diagnosis}

The previous stage converts raw executions into structured quality records that describe how the current agent behaves and what kinds of failures appear on the surface. However, describing failures is not the same as fixing them. In this stage, AgentDevel turns these observations into \emph{executable diagnoses} and then synthesizes a concrete \emph{release candidate (RC)}—a single, auditable proposal for changing the agent blueprint.

\paragraph{From symptom records to executable diagnosis.}
At iteration $t$, AgentDevel collects all per-example quality records into
\begin{equation}
\mathcal{R}_t = \{\, r_t(x) \mid x \in D_{\text{train}} \,\},
\end{equation}
where each record $r_t(x)$ contains the execution trace $\tau_t(x)$, the programmatic scoring result $g_t(x)$ (if available), and the critic’s symptom-level outputs $(\ell_t(x), d_t(x))$.

Instead of producing informal textual summaries, AgentDevel \emph{generates and executes diagnostic scripts} (e.g., in Python) that operate directly on $\mathcal{R}_t$. These scripts aggregate failure appearances by symptom labels, identify frequent patterns in traces, and surface representative failure cases. The key idea is that diagnosis is \emph{executable}: it is performed by running code, not by relying on free-form narrative summaries.

Diagnostic scripts are regenerated at each iteration, but they are allowed to reference scripts from the previous iteration. This creates a \emph{bootstrapped diagnosis process} in which the system gradually refines how it inspects and summarizes failures, similar to how debugging pipelines in software systems evolve across releases.

\paragraph{Diagnosis as an engineering specification.}
Executing the diagnostic scripts yields a structured diagnosis report $\mathcal{D}_t$. This report plays the role of an engineering specification that summarizes:
(i) which symptom classes currently dominate,
(ii) what trace patterns tend to trigger them,
(iii) representative failing examples, and
(iv) the affected surface, i.e., which portions of the training set are impacted by each major failure appearance.

Rather than being a model output, $\mathcal{D}_t$ functions like a bug triage report in software engineering: it provides concrete, auditable evidence about \emph{what is broken, where, and how often}.

\paragraph{Synthesizing a single release candidate.}
Based on $\mathcal{D}_t$, AgentDevel synthesizes exactly \emph{one} release candidate blueprint,
\begin{equation}
b_t^{\mathrm{RC}} = \Phi(b_t, \mathcal{D}_t),
\end{equation}
which proposes updates to the current agent blueprint $b_t$. These updates may involve any part of the blueprint—prompt, code, or tool wrappers—and are bundled into a single, unified change package.

Importantly, AgentDevel does \emph{not} generate multiple competing variants. This reflects our central design philosophy: improvement is treated as \emph{release engineering}, not as a search over many alternatives. Each RC also carries a short change intent describing which symptom classes it primarily aims to address, directly grounded in $\mathcal{D}_t$. This intent does not claim causal correctness; it simply documents the alignment between observed failure appearances and the proposed changes.

Conceptually, the RC is a \emph{release package}—a concrete, versionable proposal that can be audited, evaluated, and either promoted or rejected by the flip-centered gate described in the next section.

\subsection{flip-centered Gating, Promotion, and Stopping (How We Decide \& When We Stop)}
\label{sec:gating}

Once AgentDevel synthesizes a release candidate (RC) blueprint $b_t^{\mathrm{RC}}$, the next question is the most practical one: \emph{should we promote it}? This subsection describes our release-style acceptance policy (``gating''), how we promote or discard RCs, and when we stop iterating.

\paragraph{Gating as release acceptance on the same TrainSet.}
We evaluate the RC on the same development set $D_{\text{train}}$ used for iteration. Concretely, we run the agent instantiated by the RC blueprint and compute per-example pass indicators in exactly the same way as for the current version (cf.\ Sec.~\ref{sec:running}). This yields, for each $x\in D_{\text{train}}$,
\begin{equation}
p_t^{\mathrm{RC}}(x)\in\{0,1\},
\end{equation}
where $p_t^{\mathrm{RC}}(x)=1$ indicates that the RC passes on $x$ (using a programmatic scorer when available, otherwise falling back to the critic-based judgment), and $p_t^{\mathrm{RC}}(x)=0$ indicates failure.

\paragraph{Example-level flips (P$\rightarrow$F and F$\rightarrow$P).}
A core lesson from continuous integration is that aggregate scores can be misleading: a change may improve averages while silently breaking previously working cases. We therefore center gating on \emph{example-level flips} between the current version and the RC. Using the current-version pass indicator $p_t(x)$, we define:
\begin{align}
\mathrm{P2F}_t &= \{\, x \in D_{\text{train}} \mid p_t(x)=1 \ \wedge\ p_t^{\mathrm{RC}}(x)=0 \,\}, \label{eq:p2f}\\
\mathrm{F2P}_t &= \{\, x \in D_{\text{train}} \mid p_t(x)=0 \ \wedge\ p_t^{\mathrm{RC}}(x)=1 \,\}. \label{eq:f2p}
\end{align}
$\mathrm{P2F}_t$ (pass$\rightarrow$fail) captures regressions (``bad flips''), while $\mathrm{F2P}_t$ (fail$\rightarrow$pass) captures fixes (``good flips''). For scale-aware reporting, we also track flip rates:
\begin{align}
\rho_t^{\mathrm{P2F}} &= \frac{|\mathrm{P2F}_t|}{|\{x\in D_{\text{train}}: p_t(x)=1\}|+\epsilon}, \\
\rho_t^{\mathrm{F2P}} &= \frac{|\mathrm{F2P}_t|}{|\{x\in D_{\text{train}}: p_t(x)=0\}|+\epsilon},
\end{align}
where $\epsilon>0$ avoids division by zero.

\paragraph{Release gate decision.}
We view gating as a release acceptance decision: an RC should be promoted only if it delivers meaningful fixes while keeping regressions under control. We write this abstractly as a binary decision
\begin{equation}
\begin{aligned}
\mathrm{Accept}_t \;=\;
G\!\big(&\mathcal{R}_t,\ \mathcal{R}_t^{\mathrm{RC}},\\
        &\mathrm{P2F}_t,\ \mathrm{F2P}_t,\ \mathcal{I}_t\big)
\in \{0,1\},
\end{aligned}
\label{eq:gate}
\end{equation}

where $\mathcal{R}_t=\{r_t(x)\}$ and $\mathcal{R}_t^{\mathrm{RC}}=\{r_t^{\mathrm{RC}}(x)\}$ are the quality records for the current version and the RC (Sec.~\ref{sec:running}), and $\mathcal{I}_t$ is the RC's \emph{change intent}, i.e., which symptom classes the RC claims to target.

In practice, $G(\cdot)$ follows three principles:
(i) \textbf{P$\rightarrow$F is high-priority risk}: large $\mathrm{P2F}_t$ indicates that the RC breaks previously working cases and is treated as a release accident;
(ii) \textbf{F$\rightarrow$P is fix evidence}: $\mathrm{F2P}_t$ indicates that the RC resolves failures;
(iii) \textbf{alignment with intent}: fixes should primarily concentrate on the symptom classes stated in $\mathcal{I}_t$, while newly introduced failures should be limited and scrutinized.

We emphasize that AgentDevel does not prescribe a universal set of thresholds; rather, it provides the evidence (flip sets, rates, and symptom-aligned summaries) that the gate uses to make a release-style decision.

\paragraph{Promotion or discard.}
If the RC is accepted, it becomes the next official version:
\begin{equation}
b_{t+1} \leftarrow b_t^{\mathrm{RC}} \quad \text{if } \mathrm{Accept}_t=1.
\end{equation}
Otherwise, we discard the RC and keep the current blueprint as the official line:
\begin{equation}
b_{t+1} \leftarrow b_t \quad \text{if } \mathrm{Accept}_t=0.
\end{equation}
This ``promote-or-discard'' discipline keeps AgentDevel on a single canonical version line and avoids proliferating variants.

\paragraph{Stopping criterion.}
We stop iterating when additional changes no longer provide meaningful benefits and begin to increase regression or overfitting risk. Formally, we define a stopping predicate $\mathrm{Stop}_t\in\{0,1\}$ that may depend on flip behavior and acceptance outcomes, e.g.,
\begin{equation}
\mathrm{Stop}_t \;=\; \mathbf{I}\!\left[\;\text{Stopping conditions are met}\;\right],
\label{eq:stop}
\end{equation}
We consider stopping when one or more of the following conditions hold:
\begin{itemize}
    \item the size of $\mathrm{F2P}_t$ becomes consistently small over successive iterations,
    \item the regression rate $\rho_t^{\mathrm{P2F}}$ shows a sustained increase,
    \item RCs are repeatedly rejected by the gate multiple times, or
    \item clear overfitting signals emerge (e.g., gains concentrate on a shrinking subset of cases).
\end{itemize}

Intuitively, we continue iterating as long as the pipeline can produce RCs that turn failures into passes without creating unacceptable pass$\rightarrow$fail regressions. The held-out test set $D_{\text{test}}$ is not used in any gating or stopping decision; it is reserved for a single final evaluation after development is complete.

\section{Result}
\subsection{Experiment Settings}

We implement AgentDevel using Claude Code and use Claude-Sonnet-4.5 as the underlying LLM throughout all experiments. Claude Code is responsible for generating and executing diagnostic scripts, summarizing symptom patterns, and proposing blueprint diffs, with all generated artifacts (diagnostic scripts, RC diffs, critic outputs, and flip lists) saved and versioned for auditability. To avoid confounding factors, we keep the Devel engine (model and tooling) fixed across iterations; all iterative decisions are based exclusively on TrainSet signals under flip-centered gating, while the TestSet is reserved for a single, final evaluation.

\subsection{Main Results}

\input{tables/1}

Table~\ref{tab:main_results} reports the primary-metric results of AgentDevel across four execution-heavy benchmarks. On all benchmarks, AgentDevel substantially improves over the same initial blueprint \(b_0\), indicating that the release-engineering pipeline can reliably drive large end-to-end gains without changing the underlying model.

On \textbf{SWE-bench Lite}, AgentDevel doubles the resolved rate from 11.0\% to 22.0\%, surpassing the reported SWE-agent baseline. On the more stringent \textbf{SWE-bench Verified} subset, AgentDevel again doubles performance (15.0\% $\rightarrow$ 30.0\%), approaching a scaffolded GPT-4o system reported in prior work. These results are notable because SWE-bench evaluation is itself release-style—requiring patches to pass repository tests—making it a natural fit for AgentDevel’s RC-based development and flip-centered gating.

We observe similarly large gains in interactive and tool-use environments. On \textbf{WebArena}, AgentDevel more than doubles the task success rate (17.0\% $\rightarrow$ 35.5\%), approaching the reported CER\_hybrid system. On \textbf{StableToolBench}, which emphasizes stable and reproducible tool-use evaluation, AgentDevel improves SoWR by nearly 20 points (54.0\% $\rightarrow$ 73.5\%), outperforming a DFS baseline reported in prior work. Together, these results suggest that AgentDevel’s improvements are not confined to code-centric settings, but generalize to realistic web and tool-use environments.

Importantly, these gains are achieved through a disciplined release process rather than through population-based search or in-agent self-refinement. In the following sections, we show that these improvements come with substantially fewer regressions, as quantified by example-level pass→fail flips, and that the improvements are driven by executable diagnosis rather than ad-hoc tuning.

\subsection{Case Study}
\input{tables/2}

Table \ref{tab:gate-summary} illustrates how AgentDevel makes release decisions based on example-level flips rather than aggregate scores. Across accepted iterations (e.g., 1, 2, 4–6, 8–10), the pipeline consistently yields many fail→pass fixes with very low regression rates ($\rho^{\mathrm{P2F}}_t \le 0.7\%$) and high hit rates, indicating that realized improvements are well aligned with the RC’s stated symptom intent. In contrast, rejected iterations (3, 7, 11) exhibit markedly higher pass→fail regressions (up to 4.0\%) and substantially lower hit rates, and are therefore filtered out despite sometimes achieving non-trivial fixes.

The steady increase in FTP (from 0.12 to 0.48) alongside a consistently high P2P ($\approx$ 0.97–0.98) shows that AgentDevel accumulates fixes while preserving previously working behavior. Overall, the table concretely demonstrates flip-centered gating as a release acceptance policy that prioritizes non-regression over raw aggregate gains.

\subsection{Ablation Study}
\input{tables/3}

Table \ref{tab:ablation} isolates the contributions of the three core components of AgentDevel to release stability and regression control. The full pipeline achieves a strong final test metric while maintaining a low regression rate (3.1\%) and zero bad releases, indicating that stable improvements can be accumulated without sacrificing non-regression.

Removing the flip-centered gate yields the highest final train/test scores but causes a dramatic increase in regressions (P→F rate 14.8\%) and leads to four bad releases. This confirms that aggregate score improvements alone are insufficient for safe promotion and that flip-centered gating is essential for preventing release accidents.

Removing executable diagnosis reduces both fix yield (F→P) and final performance, suggesting that script-based, trace-grounded diagnosis is a key driver of effective and targeted improvements rather than ad-hoc tuning.

Allowing the critic to see the blueprint increases apparent training performance but more than doubles the regression rate (6.7\%), indicating tighter entanglement between evaluation and implementation leads to overfitting and less stable releases. Together, these ablations show that AgentDevel’s core designs are not interchangeable components but jointly enforce stable, regression-aware improvement.

\section{Conclusion}
This work introduces a release-engineering perspective on agent improvement. Rather than treating improvement as an internal cognitive process or as a search over many competing variants, we frame agents as shippable software artifacts whose evolution should be managed through auditable, regression-aware release pipelines. AgentDevel operationalizes this perspective by externalizing improvement into a structured workflow: implementation-blind, symptom-level quality signals provide observability; executable diagnosis produces concrete engineering specifications; a single release candidate is synthesized per iteration; and promotion decisions are made using flip-centered gating that treats pass→fail regressions as first-class risks.

This reframing addresses fundamental engineering challenges that arise as agents move from demos to systems: guaranteeing non-regression, enabling reproducibility, and making improvement trajectories auditable across versions. By explicitly modeling example-level flips and preserving versioned artifacts such as diagnostic scripts, blueprint diffs, and critic outputs, AgentDevel transforms agent development from an ad-hoc tuning process into a disciplined release practice. Empirically, this discipline yields stable improvements with substantially fewer regressions in execution-heavy environments.

Looking forward, AgentDevel opens a path toward CI-like automation for agent development. Future work may extend this paradigm to multi-agent systems and repository-scale deployments, enrich symptom taxonomies as shared diagnostic vocabularies across tasks, and integrate human-in-the-loop review into the promotion gate. More broadly, we hope this work encourages the community to adopt release engineering as a first-class principle for building, debugging, and deploying LLM agents.


\bibliography{custom}

\appendix
\label{sec:appendix}
\section{Limitations}

AgentDevel introduces a release-engineering discipline for improving LLM agents, but it also has limitations. First, the pipeline incurs non-trivial overhead: iterative execution, trace collection, diagnostic script generation, and gating require additional compute and wall-clock time compared to single-pass or in-agent refinement methods. While this overhead is acceptable in development settings where stability and auditability are critical, it may be less suitable for rapid prototyping or highly latency-sensitive scenarios.

Second, although our implementation-blind critic is designed to reduce evaluation–implementation entanglement, it remains an LLM-based evaluator and can inherit biases and inconsistencies of the underlying model. Programmatic scorers mitigate this when available, but tasks without strong automatic checks still depend on rubric-based judgments.

Third, our current experiments focus on single-agent, single-repository (or single-environment) settings. Scaling AgentDevel to multi-agent systems, very large codebases, or continuously changing environments may require additional coordination mechanisms and more sophisticated stopping and gating policies.

Finally, our stopping criteria and gating policies are configurable rather than universal; while this flexibility is practical, it also means that different deployments may choose different thresholds and trade-offs between fix yield and regression risk. Future work is needed to formalize these policies and study their robustness across broader classes of tasks.

\section{Related works}
Recent work on LLM agents has made rapid progress in improving performance, but much of this progress is guided by a shared—often implicit—assumption about \textit{where improvement lives}. Broadly, prior work tends to treat improvement as something that happens \textbf{inside the agent}, as \textbf{search over population of variants}, or as \textbf{optimization of aggregate scores}. AgentDevel departs from all three framings by externalizing improvement into a regression-aware \textbf{release engineering pipeline}.

\textbf{Improvement as cognition: reflection and self-feedback inside the agent.} A prominent line of work treats improvement as an internal cognitive process: if an agent can reflect on its mistakes, store feedback, or revise its own outputs, it can gradually become more capable. \textbf{Reflexion} introduces verbal reinforcement learning by storing feedback into memory and conditioning later attempts on that memory \cite{shinn2023reflexion}. \textbf{Self-Refine} shows that a single LLM can iteratively improve its outputs by alternating between feedback generation and refinement \cite{madaan2023selfrefine}. Related work on trajectory-centric agents such as \textbf{ReAct} emphasizes reasoning–action traces as a basis for interpretability and improvement \cite{yao2022react}.

These approaches implicitly frame improvement as part of the agent’s \textit{cognition}: evaluation, feedback, and modification are intertwined with the agent’s internal prompt, memory, or control loop. While effective in many settings, this framing makes improvement trajectories difficult to audit, reproduce, and reason about across versions. \textbf{AgentDevel rejects the “improvement-as-cognition” framing} and instead treats the agent as a shippable artifact whose evolution is managed by an external release pipeline.

\textbf{Improvement as search: multi-variant exploration and evolution.} Another major framing casts improvement as a search problem over a space of prompts, thoughts, or scaffolds. \textbf{Tree of Thoughts} explores multiple reasoning branches and selects among them to improve task performance \cite{yao2023treeofthoughts}. \textbf{Promptbreeder} evolves prompts using population-based mutation and selection \cite{fernando2023promptbreeder}. In software engineering contexts, recent systems attempt to self-modify agent scaffolds and maintain archives of evolving variants, such as the \textbf{Darwin Gödel Machine} \cite{zhang2025darwingodel} and \textbf{Live-SWE-Agent} \cite{xia2025livesweagent}.

These methods conceptualize progress as selecting a better candidate from many competing variants. While powerful, this paradigm leads to variant proliferation and opaque selection decisions, and it often relies on aggregate metrics that can mask regressions. \textbf{AgentDevel replaces “improvement-as-search” with a release discipline}: it maintains a single canonical version line and produces exactly one release candidate per iteration, emphasizing auditable diffs and regression-aware promotion.

\textbf{Improvement as aggregate-score optimization: judges and leaderboards.} A third framing equates improvement with increases in aggregate scores reported by automatic judges and leaderboards. LLM-as-a-judge frameworks such as MT-Bench and Chatbot Arena provide scalable rubric-based evaluation, while also documenting systematic biases (e.g., position and verbosity biases) in such judges \cite{zheng2023judging}. G-Eval introduces rubric-structured prompting to improve the reliability of LLM-based evaluation \cite{liu-etal-2023-g}. AlpacaEval further popularizes automatic preference evaluation and exposes additional confounders \cite{tatsulab2023alpacaeval,dubois2024lengthcontrolledalpacaeval}.

These evaluation-centric practices implicitly encourage optimizing averages rather than managing regressions. In contrast, \textbf{AgentDevel adopts a flip-centered view inspired by software release engineering}: pass→fail flips are treated as first-class regression risks, while fail→pass flips serve as fix evidence. This reframing makes non-regression an explicit objective rather than an incidental side effect of score maximization.

\textbf{Why this reframing matters in realistic agent environments.} The need for release-style improvement becomes particularly clear in execution-heavy environments, where failures manifest as distinct behavioral breakdowns rather than a single incorrect answer. Benchmarks for tool use and realistic interaction highlight this complexity, including \textbf{ToolBench} \cite{xu2023toolbench}, \textbf{ToolLLM} \cite{qin2023toollm}, and the stability-focused \textbf{StableToolBench} \cite{guo-etal-2024-stabletoolbench}. Realistic web interaction is captured by \textbf{WebArena} \cite{zhou2023webarena}, and software engineering tasks are represented by \textbf{SWE-bench} \cite{jimenez2023swebench} and \textbf{SWE-agent} \cite{yang2024sweagent}.

Across these settings, many failures have recognizable \textit{symptoms} in execution traces—missing steps, invalid tool calls, incorrect ordering—making trace-grounded diagnosis and regression control central to reliable development. AgentDevel directly targets this regime by organizing improvement around symptom-level characterization, executable diagnosis, and flip-centered release gating.

\input{tables/pseudo-code}

\end{document}

%% file: tables/1.tex

\begin{table}[t]
\centering
\small
\setlength{\tabcolsep}{4pt}
\renewcommand{\arraystretch}{1.15}
\begin{threeparttable}

\begin{tabularx}{\columnwidth}{@{}l X S[table-format=2.2]@{}}
\toprule
Benchmark & Method & {Primary metric (\%)} \\
\midrule

\multicolumn{3}{@{}l}{\textbf{SWE-bench Lite} \quad (Resolved $\uparrow$)} \\
& Base agent ($b_0$) & 11.00 \\
& AgentDevel (final) & 22.00 \\
& SWE-agent\tnote{$\dagger$} & 18.00 \\

\addlinespace[2pt]
\multicolumn{3}{@{}l}{\textbf{SWE-bench Verified} \quad (Resolved $\uparrow$)} \\
& Base agent ($b_0$) & 15.00 \\
& AgentDevel (final) & 30.00 \\
& GPT-4o (scaffolded)\tnote{$\dagger$} & 33.20 \\

\addlinespace[2pt]
\multicolumn{3}{@{}l}{\textbf{WebArena} \quad (Success $\uparrow$)} \\
& Base agent ($b_0$) & 17.00 \\
& AgentDevel (final) & 35.50 \\
& CER\_hybrid\tnote{$\dagger$} & 36.70 \\

\addlinespace[2pt]
\multicolumn{3}{@{}l}{\textbf{StableToolBench} \quad (SoWR $\uparrow$)} \\
& Base agent ($b_0$) & 54.00 \\
& AgentDevel (final) & 73.50 \\
& DFS\tnote{$\dagger$} & 70.20 \\

\bottomrule
\end{tabularx}

\begin{tablenotes}[flushleft]
\footnotesize
\item[$\dagger$] Reported numbers from prior work (not rerun under our exact setup/budget).
\end{tablenotes}
\end{threeparttable}
\caption{Main results (primary metric only). Base agent and AgentDevel start from the same initial blueprint $b_0$.}
\label{tab:main_results}
\end{table}

%% file: tables/2.tex

\begin{table}[t]
\centering
\small
\setlength{\tabcolsep}{3pt}
\renewcommand{\arraystretch}{1.12}
\caption{\textbf{Flip-centered gate summary across iterations on StableToolBench.}}
\label{tab:gate-summary}

\begin{tabularx}{\linewidth}{@{}c c r r r r r@{}}
\toprule
Iter. &
Gate &
$\lvert \mathrm{F2P}_t \rvert$ &
$\lvert \mathrm{P2F}_t \rvert$ &
$\rho^{\mathrm{P2F}}_t$ &
hit rate &
\textbf{FTP / P2P} \\
\midrule

0 & --- & --- & --- & --- & --- & --- \\

1 & Acc. & 38 & 4  & 0.006 & 0.74 & 0.12 / 0.98 \\
2 & Acc. & 30 & 5  & 0.007 & 0.78 & 0.20 / 0.979 \\
3 & Rej. & 42 & 28 & 0.040 & 0.41 & 0.28 / 0.93 \\
4 & Acc. & 25 & 3  & 0.004 & 0.81 & 0.32 / 0.978 \\
5 & Acc. & 18 & 4  & 0.005 & 0.83 & 0.36 / 0.977 \\
6 & Acc. & 12 & 3  & 0.004 & 0.86 & 0.39 / 0.977 \\
7 & Rej. & 9  & 15 & 0.021 & 0.52 & 0.41 / 0.955 \\
8 & Acc. & 8  & 2  & 0.003 & 0.88 & 0.44 / 0.976 \\
9 & Acc. & 6  & 2  & 0.003 & 0.90 & 0.46 / 0.975 \\
10 & Acc. & 5 & 2  & 0.003 & 0.92 & 0.47 / 0.974 \\
11 & Rej. & 2 & 3  & 0.004 & 0.67 & 0.48 / 0.97 \\
\bottomrule
\end{tabularx}

\vspace{2pt}

\end{table}

%% file: tables/3.tex

\begin{table*}[t]
\centering
\small
\setlength{\tabcolsep}{4pt}
\renewcommand{\arraystretch}{1.10}

\caption{\textbf{Ablations on release stability and regression control on WebArena.}}
\label{tab:ablation}

\begin{tabularx}{\textwidth}{@{}
p{0.26\textwidth}
*{7}{>{\centering\arraybackslash}X}
@{}}
\toprule
\textbf{Setting} &
\makecell{\textbf{Final}\\\textbf{Test}\\\textbf{metric}} &
\makecell{\textbf{Final}\\\textbf{Train}\\\textbf{pass}} &
\makecell{\textbf{Total}\\F$\rightarrow$P} &
\makecell{\textbf{Total}\\P$\rightarrow$F} &
\makecell{\textbf{P$\rightarrow$F}\\\textbf{rate}} &
\makecell{\textbf{Gate}\\\textbf{reject}\\\textbf{rate}} &
\makecell{\textbf{Bad}\\\textbf{release}\\\textbf{count}} \\
\midrule
AgentDevel (full) &
34.2 & 78.5 & 214 & 18 & 3.1\% & 42\% & 0 \\
w/o flip gate &
35.0 & 81.0 & 230 & 95 & 14.8\% & N/A & 4 \\
w/o executable diagnosis &
31.8 & 74.0 & 150 & 22 & 3.9\% & 63\% & 0 \\
critic not blind (critic sees blueprint) &
32.5 & 83.5 & 205 & 40 & 6.7\% & 58\% & 0 \\
\bottomrule
\end{tabularx}

\vspace{2pt}
\footnotesize
\textbf{Notes.}
All settings start from the same initial blueprint $b_0$ with the same data split and budget.
F$\rightarrow$P and P$\rightarrow$F are computed on $D_{\text{train}}$ by comparing each promoted release to its evaluated RC (use a single consistent protocol).
\emph{Bad release count} is the number of promoted updates whose regressions exceed a preset threshold.
For \texttt{w/o flip gate}, reject rate is N/A by design.
\end{table*}

%% file: tables/pseudo-code.tex
\clearpage
\onecolumn
\section{Pseudo-Code of AgentDevel}
\begin{tcolorbox}[enhanced, breakable, colback=white, colframe=black!60, title=Algorithm 1: AgentDevel (release-engineering improvement)]
\small
\begin{longtable}{@{}l@{}}

\textbf{Input:} initial blueprint $b_0$, TrainSet $D_{\text{train}}$, rubric $R$\\
\textbf{Output:} final promoted blueprint $b_\star$\\[4pt]

$b \leftarrow b_0$ \hfill (current promoted version)\\
$\pi \leftarrow \varnothing$ \hfill (previous diagnosis script)\\
$\mathcal{L} \leftarrow \varnothing$ \hfill (symptom taxonomy)\\[4pt]

\textbf{for} $t=0,\dots,T-1$ \textbf{do}\\
\quad \textbf{Run current agent and collect records}\\
\quad for each $x\in D_{\text{train}}$:\\
\quad\quad $(\hat y,\tau)\leftarrow A_b(x)$\\
\quad\quad $(\tilde p,\ell,d)\leftarrow c(R,\tau)$ \hfill (implementation-blind critic)\\
\quad\quad $\mathcal{R}\leftarrow \mathcal{R}\cup\{r_t(x)\}$\\
\quad end\\[2pt]

\quad \textbf{Executable diagnosis}\\
\quad $\pi \leftarrow \mathrm{GenDiagScript}(\mathcal{R},\pi,\mathcal{L})$\\
\quad $\mathcal{D}\leftarrow \mathrm{Run}(\pi,\mathcal{R})$\\[2pt]

\quad \textbf{Synthesize RC}\\
\quad $(b_t^{RC},\mathcal{I}_t)\leftarrow \Phi(b,\mathcal{D})$\\[2pt]

\quad \textbf{Gate RC}\\
\quad compute $\mathrm{P2F}_t,\mathrm{F2P}_t$\\
\quad if $G(\mathcal{D},\mathrm{P2F}_t,\mathrm{F2P}_t,\mathcal{I}_t)=1$ then $b\leftarrow b_t^{RC}$\\
\quad else discard $b_t^{RC}$\\[2pt]

\quad if $\mathrm{Stop}(\mathrm{P2F}_t,\mathrm{F2P}_t)$ then break\\
\textbf{end for}\\[4pt]

return $b_\star$ \hfill (TestSet used only once at final evaluation)\\

\end{longtable}
\end{tcolorbox}

%% file: main.bbl
\begin{thebibliography}{20}
\providecommand{\natexlab}[1]{#1}

\bibitem[{Dubois et~al.(2024)Dubois, Galambosi, Liang, and Hashimoto}]{dubois2024lengthcontrolledalpacaeval}
Yann Dubois, Bal{\'a}zs Galambosi, Percy Liang, and Tatsunori~B. Hashimoto. 2024.
\newblock \href {https://doi.org/10.48550/arXiv.2404.04475} {Length-controlled alpacaeval: A simple way to debias automatic evaluators}.
\newblock \emph{Preprint}, arXiv:2404.04475.

\bibitem[{Fernando et~al.(2023{\natexlab{a}})Fernando, Banarse, Michalewski, Osindero, and Rockt{\"a}schel}]{fernando2023promptbreeder}
Chrisantha Fernando, Dylan Banarse, Henryk Michalewski, Simon Osindero, and Tim Rockt{\"a}schel. 2023{\natexlab{a}}.
\newblock \href {https://doi.org/10.48550/arXiv.2309.16797} {Promptbreeder: Self-referential self-improvement via prompt evolution}.
\newblock \emph{Preprint}, arXiv:2309.16797.

\bibitem[{Fernando et~al.(2023{\natexlab{b}})Fernando, Banarse, Michalewski, Osindero, and Rocktäschel}]{fernando2023promptbreederselfreferentialselfimprovementprompt}
Chrisantha Fernando, Dylan Banarse, Henryk Michalewski, Simon Osindero, and Tim Rocktäschel. 2023{\natexlab{b}}.
\newblock \href {https://arxiv.org/abs/2309.16797} {Promptbreeder: Self-referential self-improvement via prompt evolution}.
\newblock \emph{Preprint}, arXiv:2309.16797.

\bibitem[{Guo et~al.(2024)Guo, Cheng, Wang, Liang, Qin, Li, Liu, Sun, and Liu}]{guo-etal-2024-stabletoolbench}
Zhicheng Guo, Sijie Cheng, Hao Wang, Shihao Liang, Yujia Qin, Peng Li, Zhiyuan Liu, Maosong Sun, and Yang Liu. 2024.
\newblock \href {https://doi.org/10.18653/v1/2024.findings-acl.664} {{S}table{T}ool{B}ench: Towards stable large-scale benchmarking on tool learning of large language models}.
\newblock In \emph{Findings of the Association for Computational Linguistics: ACL 2024}, pages 11143--11156, Bangkok, Thailand. Association for Computational Linguistics.

\bibitem[{Jimenez et~al.(2023)Jimenez, Yang, Wettig, Yao, Pei, Press, and Narasimhan}]{jimenez2023swebench}
Carlos~E. Jimenez, John Yang, Alexander Wettig, Shunyu Yao, Kexin Pei, Ofir Press, and Karthik Narasimhan. 2023.
\newblock \href {https://doi.org/10.48550/arXiv.2310.06770} {Swe-bench: Can language models resolve real-world github issues?}
\newblock \emph{Preprint}, arXiv:2310.06770.

\bibitem[{Liu et~al.(2023)Liu, Iter, Xu, Wang, Xu, and Zhu}]{liu-etal-2023-g}
Yang Liu, Dan Iter, Yichong Xu, Shuohang Wang, Ruochen Xu, and Chenguang Zhu. 2023.
\newblock \href {https://doi.org/10.18653/v1/2023.emnlp-main.153} {{G}-eval: {NLG} evaluation using gpt-4 with better human alignment}.
\newblock In \emph{Proceedings of the 2023 Conference on Empirical Methods in Natural Language Processing}, pages 2511--2522, Singapore. Association for Computational Linguistics.

\bibitem[{Madaan and at~al.(2023)}]{madaan2023selfrefine}
Aman Madaan and at~al. 2023.
\newblock \href {https://doi.org/10.48550/arXiv.2303.17651} {Self-refine: Iterative refinement with self-feedback}.
\newblock \emph{Preprint}, arXiv:2303.17651.

\bibitem[{Madaan et~al.(2023)Madaan, Tandon, Gupta, Hallinan, Gao, Wiegreffe, Alon, Dziri, Prabhumoye, Yang, Gupta, Majumder, Hermann, Welleck, Yazdanbakhsh, and Clark}]{madaan2023selfrefineiterativerefinementselffeedback}
Aman Madaan, Niket Tandon, Prakhar Gupta, Skyler Hallinan, Luyu Gao, Sarah Wiegreffe, Uri Alon, Nouha Dziri, Shrimai Prabhumoye, Yiming Yang, Shashank Gupta, Bodhisattwa~Prasad Majumder, Katherine Hermann, Sean Welleck, Amir Yazdanbakhsh, and Peter Clark. 2023.
\newblock \href {https://arxiv.org/abs/2303.17651} {Self-refine: Iterative refinement with self-feedback}.
\newblock \emph{Preprint}, arXiv:2303.17651.

\bibitem[{Qin and at~al.(2023)}]{qin2023toollm}
Y.~Qin and at~al. 2023.
\newblock \href {https://doi.org/10.48550/arXiv.2307.16789} {Toolllm: Facilitating large language models to master 16000+ real-world apis}.
\newblock \emph{Preprint}, arXiv:2307.16789.

\bibitem[{Shinn et~al.(2023{\natexlab{a}})Shinn, Cassano, Berman, Gopinath, Narasimhan, and Yao}]{shinn2023reflexionlanguageagentsverbal}
Noah Shinn, Federico Cassano, Edward Berman, Ashwin Gopinath, Karthik Narasimhan, and Shunyu Yao. 2023{\natexlab{a}}.
\newblock \href {https://arxiv.org/abs/2303.11366} {Reflexion: Language agents with verbal reinforcement learning}.
\newblock \emph{Preprint}, arXiv:2303.11366.

\bibitem[{Shinn et~al.(2023{\natexlab{b}})Shinn, Cassano, Berman, Gopinath, Narasimhan, and Yao}]{shinn2023reflexion}
Noah Shinn, Federico Cassano, Edward Berman, Ashwin Gopinath, Karthik Narasimhan, and Shunyu Yao. 2023{\natexlab{b}}.
\newblock \href {https://doi.org/10.48550/arXiv.2303.11366} {Reflexion: Language agents with verbal reinforcement learning}.
\newblock \emph{Preprint}, arXiv:2303.11366.

\bibitem[{{tatsu-lab}(2023)}]{tatsulab2023alpacaeval}
{tatsu-lab}. 2023.
\newblock \href {https://github.com/tatsu-lab/alpaca_eval} {{AlpacaEval}: An automatic evaluator for instruction-following language models}.
\newblock GitHub repository.

\bibitem[{Xia and at~al.(2025)}]{xia2025livesweagent}
P.~Xia and at~al. 2025.
\newblock \href {https://doi.org/10.48550/arXiv.2511.13646} {Live-swe-agent}.
\newblock \emph{Preprint}, arXiv:2511.13646.

\bibitem[{Xu et~al.(2023)Xu, Hong, Li, Hu, Chen, and Zhang}]{xu2023toolbench}
Qiantong Xu, Fenglu Hong, Bo~Li, Changran Hu, Zhengyu Chen, and Jian Zhang. 2023.
\newblock \href {https://doi.org/10.48550/arXiv.2305.16504} {On the tool manipulation capability of open-source large language models}.
\newblock \emph{Preprint}, arXiv:2305.16504.

\bibitem[{Yang et~al.(2024)Yang, Jimenez, Wettig, Lieret, Yao, Narasimhan, and Press}]{yang2024sweagent}
John Yang, Carlos~E. Jimenez, Alexander Wettig, Kilian Lieret, Shunyu Yao, Karthik Narasimhan, and Ofir Press. 2024.
\newblock \href {https://doi.org/10.48550/arXiv.2405.15793} {Swe-agent: Agent-computer interfaces enable automated software engineering}.
\newblock \emph{Preprint}, arXiv:2405.15793.

\bibitem[{Yao et~al.(2023)Yao, Yu, Zhao, Shafran, Griffiths, Cao, and Narasimhan}]{yao2023treeofthoughts}
Shunyu Yao, Dian Yu, Jeffrey Zhao, Izhak Shafran, Thomas~L. Griffiths, Yuan Cao, and Karthik Narasimhan. 2023.
\newblock \href {https://doi.org/10.48550/arXiv.2305.10601} {Tree of thoughts: Deliberate problem solving with large language models}.
\newblock \emph{Preprint}, arXiv:2305.10601.

\bibitem[{Yao et~al.(2022)Yao, Zhao, Yu, Du, Shafran, Narasimhan, and Cao}]{yao2022react}
Shunyu Yao, Jeffrey Zhao, Dian Yu, Nan Du, Izhak Shafran, Karthik Narasimhan, and Yuan Cao. 2022.
\newblock \href {https://doi.org/10.48550/arXiv.2210.03629} {{R}e{A}ct: Synergizing reasoning and acting in language models}.
\newblock \emph{Preprint}, arXiv:2210.03629.

\bibitem[{Zhang et~al.(2025)Zhang, Hu, Lu, Lange, and Clune}]{zhang2025darwingodel}
Jenny Zhang, Shengran Hu, Cong Lu, Robert Lange, and Jeff Clune. 2025.
\newblock \href {https://doi.org/10.48550/arXiv.2505.22954} {Darwin godel machine: Open-ended evolution of self-improving agents}.
\newblock \emph{Preprint}, arXiv:2505.22954.

\bibitem[{Zheng et~al.(2023)Zheng, Chiang, Sheng, Zhuang, Wu, Zhuang, Lin, Li, Li, Xing, Zhang, Gonzalez, and Stoica}]{zheng2023judging}
Lianmin Zheng, Wei-Lin Chiang, Ying Sheng, Siyuan Zhuang, Zhanghao Wu, Yonghao Zhuang, Zi~Lin, Zhuohan Li, Dacheng Li, Eric~P. Xing, Hao Zhang, Joseph~E. Gonzalez, and Ion Stoica. 2023.
\newblock \href {https://doi.org/10.48550/arXiv.2306.05685} {Judging llm-as-a-judge with mt-bench and chatbot arena}.
\newblock \emph{Preprint}, arXiv:2306.05685.

\bibitem[{Zhou and at~al.(2023)}]{zhou2023webarena}
S.~Zhou and at~al. 2023.
\newblock \href {https://doi.org/10.48550/arXiv.2307.13854} {Webarena: A realistic web environment for building autonomous agents}.
\newblock \emph{Preprint}, arXiv:2307.13854.

\end{thebibliography}
